\documentclass[runningheads]{llncs}
\usepackage{graphicx}
\usepackage{amsmath,amssymb} 
\usepackage{color}
\usepackage{hyperref}

\begin{document}

\title{Knowledge Distillation with Feature Maps for Image Classification} 
\titlerunning{KDFM} 


\author{Wei-Chun Chen \and
Chia-Che Chang \and Chien-Yu Lu \and Che-Rung Lee}
%

\authorrunning{W. Chen et al.} 


\institute{National Tsing Hua University, Taiwan\\ \email{\{meatybobby,chang810249,j19550713\}@gmail.com,\\cherung@cs.nthu.edu.tw}}

\maketitle

\begin{abstract}
The model reduction problem that eases the computation costs and latency of complex deep learning architectures has received an increasing number of investigations owing to its importance in model deployment. One promising method is knowledge distillation (KD), which creates a fast-to-execute student model to mimic a large teacher network.   In this paper, we propose a method, called KDFM (Knowledge Distillation with Feature Maps), which improves the effectiveness of KD by learning the feature maps from the teacher network. Two major techniques used in KDFM are shared classifier and generative adversarial network.  Experimental results show that KDFM can use a four layers CNN to mimic DenseNet-40 and use MobileNet to mimic DenseNet-100. Both student networks have less than 1\% accuracy loss comparing to their teacher models for CIFAR-100 datasets. The student networks are 2-6 times faster than their teacher models for inference, and the model size of MobileNet is less than half of DenseNet-100's.

\keywords{Knowledge Distillation  \and Model Compression \and Generative Adversarial Network.}
\end{abstract}
\section{Introduction}
Deep learning has shown its capability of solving various computer vision problems, such as image classification \cite{Krizhevsky:2012:ICD:2999134.2999257} and object detection \cite{DBLP:journals/corr/GirshickDDM13}.  Its success also enables many related applications, such as self-driving cars \cite{DBLP:journals/corr/BojarskiTDFFGJM16}, medical diagnosis \cite{6467091}, and intelligent manufacturing \cite{intelligent-manu}.  

However, the state-of-the-art deep learning models usually have large memory footprints and require intensive computational power. For instance, VGGNet \cite{DBLP:journals/corr/SimonyanZ14a} requires more than 100 million parameters and more than 15 giga floating-point-operations (GFLOPs) to inference an image of $224 \times 224$ resolution. It is difficult to deploy these models on some platforms with limited resources, such as mobile devices, or Internet of Things (IOT) devices. In addition, the inference time may be too long to satisfy the real-time requests of tasks.

Many methods have been proposed to reduce the computational costs of deep learning models during the inference time. For instance, the weight quantization method \cite{DBLP:journals/corr/HubaraCSEB16}  reduces the network size by quantizing the network parameters. Structure pruning \cite{NIPS1992_647} is another example that removes the unnecessary parameters or channels of a trained convolutional neural network (CNN), and then fine-tunes the model to gain higher accuracy. These methods have achieved competitive accuracy with less model size comparing to those of original models.  Although they can effectively reduce the model sizes and inference time, their operations are usually not matching the instructions of commodity acceleration hardware, such as GPU or TPU.  As a result, the real performance gain of those methods may not be significant comparing to the original models with hardware acceleration.

Another promising directions of model reduction is Knowledge Distillation (KD) \cite{2015arXiv150302531H}, whose idea is to train a student network to mimic the ability of a teacher model.  The student model is usually smaller or faster-to-execute than the teacher model. Hinton \& Dean \cite{2015arXiv150302531H} coined the name of Knowledge Distillation (KD).  They trained student networks by the ``soft target'', a modify softmax function which can provide more information than the traditional softmax function. The experiment shows KD can improve the performance of a single shallow network by distilling the knowledge in an ensemble model. Romero \& Bengio \cite{Romero15-iclr} extended the idea of KD and proposed FITNET. They trained thinner and deeper student networks by the ``intermediate-level hint'', which is from the hidden layers of the teacher network, and the ``soft target'' to learn the teacher network. The results show that FITNET can use fewer parameters to mimic the teacher network. In \cite{iclr2018training}, Xu \& Huang  proposed the method that uses  conditional adversarial networks to make student networks learn the logits of the teacher networks. Their experiments showed that it can further improve the performance of student models trained by traditional KD.

However, those KD methods only learn the logits of teacher models. They are usually not powerful enough to make student models mimic all kinds of teacher models well. They often need to customize the student models for specific architectures. In addition, as the deep models become more and more complicated, the effectiveness of previous methods for knowledge distillation decreases.  One example is DenseNet \cite{8099726}, which connects all layers directly with each other, and requires more computation in inference time.  In our experiments, the simple CNN student models learned from previous methods cannot achieve the similar accuracy as the teacher model.

In this paper, we propose a method, called KDFM (Knowledge Distillation with Feature Maps), which learns the feature maps from the teacher model.  For the application of image classification, feature maps often provide more information than logits.  The feature maps in the last layer are used because they possess the high level features of the input images, which are the most informative for classification. KDFM utilizes two techniques to distill the knowledge of feature maps.  First, it lets the teacher model and the student model share the classifier.  Through the training of the shared classifier, the student model can learn the feature maps from the teacher.  Second, the idea of generative adversarial networks (GANs) \cite{NIPS2014_5423} is used to improve the learning process.  The feature map in CNN is a special type of images.  During the learning process, the discriminator is forcing the student model (generator) to generate similar feature maps to those of the teacher model (inputs of GANs).

Although the method could be generally applied to other types of networks, we employ the DenseNets as the teacher models to illustrate the idea and to demonstrate its effectiveness in the experiments.  Unlike FITNET \cite{Romero15-iclr} whose student models are thin and deep, we let the student models be shallow and fat, because such kind of networks are easier to be parallelized on modern accelerators, such as GPU.

We validated the effectiveness of KDFM using CIFAR-100 datasets \cite{cifar100} and ImageNet datasets \cite{imagenet_cvpr09}. The first experiment uses a simple student network which only contains 4 convolutional layers and a fully-connected layer to mimic DenseNet-40 (DenseNet with 40 layers) on CIFAR-100.  The result shows the student model generated by KDFM has less than 1\% accuracy loss and 2 times faster inference time comparing to DenseNet-40, which is better than other methods.  The second experiment trains the model of MobileNet \cite{DBLP:journals/corr/HowardZCKWWAA17}, a state-of-the-art network for mobile and embedded platforms, to mimic DenseNet-100 (DenseNet with 100 layers) on CIFAR-100.  The results show that the student model is more than 6 times faster than DenseNet-100 in terms of inference time, with only half model size and less than 1\% accuracy loss. The third experiment uses MobileNet v2 \cite{DBLP:MBv2} to mimic ResNet-152 \cite{DBLP:resnet} on ImageNet, and the accuracy of KDFM is better than other KD methods.

The rest of paper is organized as follows.  Section \ref{sec:related} gives a brief illustration of knowledge distillation (KD) and generative adversarial networks (GANs).  Section \ref{sec:model} introduces the design of KDFM to construct a student model.  Section \ref{sec:exp} shows the experimental results and the performance comparison with other methods.  The conclusion and future work are presented in the last section.

\section{Related Work}
\label{sec:related}

\subsection{Knowledge Distillation}
In \cite{NIPS2014_5484}, Ba \& Caruna asked an interesting question, ``Do Deep Nets Really Need to be Deep?''  Their answer is that shallow nets can be trained to perform similarly to complex, well-engineered, deeper convolutional models.  The method they used to train shallow networks is mimicking the teacher networks' logits, the value before the softmax activation. In 2017, authors presented more experimental results in \cite{2016arXiv160305691U}.

Hinton \& Dean \cite{2015arXiv150302531H} generalized this idea as Knowledge Distillation (KD).  The concept of knowledge distillation is to train a student network by a hard target $P_{H}$ and a soft target $P_{S}$:
\begin{align}
& P_{H}(x)=softmax(x) \label{eqn:PH}\\
& P_{S}(x,t)=softmax\left(\frac{x}{t}\right) \label{eqn:PS}
\end{align}
where $x$ are logits in a neural network, and $t$ is a hyper-parameter, $t>1$, to soften the probability distribution over classes. A higher value of $t$ could provide more information.

Let $x_{T}$ be the logits of the teacher network and $x_{S}$ be the logits of the student network.  The goal of student network is to optimize the loss function

\begin{equation}
    L_{KD}=\lambda L_{H}+(1-\lambda)L_{S},
\end{equation}
where
\begin{align}
\label{eqn:HKD}
    & L_{H}=\boldsymbol{H}(P_{H}(x_{S}),y)) \mbox{ and }\\
    & L_{S}=\boldsymbol{H}(P_{S}(x_{S},t), P_{S}(x_{T},t))
\end{align}
and $y$ is ground-truth label.
They trained shallow networks by the ``soft target'' of teacher networks. KD softens the output of the softmax function, providing more information than traditional softmax functions. The experiment in this paper shows KD can improve the performance of a model by distilling the knowledge in an ensemble model into a single model.

Romero \& Bengio \cite{Romero15-iclr} proposed FITNET, which extends the idea of KD by using ``intermediate-level hints'' from the hidden layers of the teacher network to guide the student networks. They train thinner and deeper student networks to learn the intermediate representations and the soft target of the teacher network. The results show that the student network of FITNET can perform comparable or even better than the teacher network with fewer parameters.

\subsection{Generative Adversarial Networks}
Generative Adversarial Networks (GANs) have shown impressive results for unsupervised learning tasks, such as image generation \cite{NIPS2014_5423}, image synthesis \cite{pmlr-v48-reed16}, and image super-resolution \cite{8099502}. A GAN usually consists of two modules: a generator (G) and a discriminator (D). In a typical GAN model, the discriminator learns to distinguish real samples and fake results produced by the generator, and the generator learns to create samples which can be judged as real ones by the discriminator.

Mirza \& Osindero \cite{DBLP:journals/corr/MirzaO14} extended GANs to a conditional model by feeding extra information, such as class labels, to the generator and discriminator. Chen \& Abbeel \cite{DBLP:journals/corr/ChenDHSSA16} proposed InfoGAN, an information-theoretic extension to GANs, which is able to learn disentangle representation. Some studies \cite{2016arXiv160601583O,NIPS2016_6125,pmlr-v70-odena17a} modify the discriminator to contain an auxiliary decoder network that can output class labels for training data.

\subsection{DenseNet}
Huang \& Weinberger \cite{8099726} proposed a new architecture, DenseNet, which connects all layers directly with each other. This idea is extended from ResNet \cite{7780459} which aggregates previous feature maps and feeds the summation into a layer. Different from ResNet, DenseNet concatenates the feature maps from all preceding layers. It requires fewer parameters than traditional convolutional networks, because it doesn't need to relearn redundant feature maps. It performs state-of-the-art results on most classification benchmark tasks.

\subsection{MobileNet}
Howard \& Kalenichenko \cite{DBLP:journals/corr/HowardZCKWWAA17} proposed MobileNet for mobile and embedded platforms. MobileNet uses depth-wise separable convolutions to reduce the computation and build a light-weight network. MobileNet allows to build the model on resource and accuracy trade-offs by using width multiplier and resolution multiplier. The effectiveness of MobileNet has been  demonstrated across a wide range of applications.

\section{The Design of KDFM}
\label{sec:model}
KDFM use a GAN with an auxiliary decoder network that can output class labels for training data.  More specifically, it consists of three components, a generator $G$, a discriminator $D$, and a classifier $C$.  The generator $G$ is a feature extractor who produces the feature maps from the input images. The discriminator $D$ distinguishes the real feature map, generated by the teacher network, and the fake feature map, generated by $G$.  The classifier $C$ is a feature decoder, whose inputs are also feature maps, and outputs are the hard target and the soft target, as defined in (\ref{eqn:PH}) and (\ref{eqn:PS}). 

The goal of KDFM is to make $G$ learn the feature map from the teacher network, and to train $C$ to classify the images based on the feature maps.  Two objective functions, adversarial loss and knowledge distillation loss, are designed to achieve the goal.  The adversarial loss of KDFM is adopted from the objective function of LSGAN \cite{8237566},
\begin{align}
& L_{advD}=\frac{1}{2}[D(G(X))]^{2}+\frac{1}{2}[D(T(X))-1]^{2}\label{eqn:advD} \\
& L_{advG}=\frac{1}{2}[D(G(X))-1]^{2} \label{eqn:advG}
\end{align}
where $X$ denotes the input images, $G(X)$ is the feature maps generated by $G$, and $T(X)$ is the feature maps generated by the teacher model.  The function $D$ is designed to discriminate between the real feature map $T(X)$  and the fake feature map $G(X)$. We chose LSGAN because it is the state-of-the-art GAN model and the range of its loss function can be easily combined with the knowledge distillation loss.

The knowledge distillation loss in KDFM is defined as below:
\begin{equation}
    \label{eqn:KD}
    L_{KD}=\lambda L_{H}+(1-\lambda)L_{S}
\end{equation}
where
\begin{align}
\label{eqn:HKDFM}
& L_{H}=\boldsymbol{H}(P_{H}(C(G(X))),P_{H}(z))+\boldsymbol{H}(P_{H}(C(T(X))), P_{H}(z))\\
& L_{S}=\boldsymbol{H}(P_{S}(C(G(X)),t), P_{S}(z,t))+\boldsymbol{H}(P_{S}(C(T(X)),t), P_{S}(z,t))
\end{align}
the value $z$ is the logits from the teacher network, $\boldsymbol{H}$ refers to cross-entropy, and $\lambda$ is a hyper-parameter, $0<\lambda< 1$, controlling the ratio of $L_{H}$ and $L_{S}$.  If the student model is similar to the teacher model, $\lambda$ need not be large.  Besides, we change the ground-truth label to the label of the teacher network in (\ref{eqn:HKDFM}). The experiment also shows that it achieves better accuracy.

Unlike traditional GAN, the loss function of $G$ in KDFM combines the adversarial loss and the knowledge distillation loss, 
\begin{align}
\label{eqn:G loss}
L_{G}=L_{advG}+\alpha L_{KD}
\end{align}
where $\alpha$ is a hyper-parameter to balance the scale of the adversarial loss and the knowledge distillation loss.

The training of KDFM is to minimize the loss functions of three components simultaneously.  For the generator $G$,  the loss function is $L_G$, as defined in (\ref{eqn:G loss}); for the discriminator $D$, the loss function is $L_{advD}$, as defined in (\ref{eqn:advD}); and for the classifier $C$, the loss function is $L_{KD}$, defined in (\ref{eqn:KD}).

\begin{figure}
\centering
\includegraphics[width=10cm]{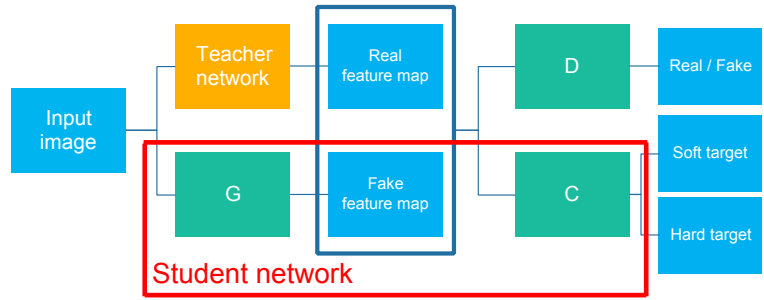}
\caption{Overview of KDFM, consisting of three module, a discriminator D, a generator G, and a classifier C. G and C compose a student network. The student network outputs the hard target for the inference.}
\label{fig:kdgan_arch}
\end{figure}

Figure \ref{fig:kdgan_arch} shows the network architecture of KDFM.  The student network consists of two parts, the feature extractor $G$ and the feature decoder $C$. The feature extractor generates the feature map, and the feature decoder classifies the feature map to probability distribution over classes.  After each components are well-trained, the student network is constructed from $G$ and $C$.  In our design, $C$ only has a pooling layer and one fully connected layer. 

The training process works like the alternative least square (ALS) method.  Let's use $L_H$ to illustrate the idea, since $L_S$ has the same structure.  To minimize  $\boldsymbol{H}(P_{H}(C(T(X))), P_{H}(z))$, the classifier $C$ needs to learn teacher network's hard target $P_{H}(z)$.  Meanwhile, the term $L_{KD}$ is also added to the loss function of the student network.  To minimize the  $\boldsymbol{H}(P_{H}(C(G(X))),P_{H}(z))$, the student model must output feature maps $G(X)$ similar to $T(X)$, so that $P_H(C(G(X)))$ can approximate $P_H(z)$. 

\section{Experiments}
\label{sec:exp}
We validated the effectiveness of KDFM using CIFAR-100 and ImageNet datasets.  We used DenseNet and ResNet as the teacher models, whose implementations \cite{vision_networks} are in TensorFlow, and followed the standard training process with data augmentation. Two types of student models are used in the experiments.  The first kind of student models are simple convolutional neural networks (CNNs) that consist of several convolutional layers and one fully-connected layer, with ReLU activation \cite{Nair:2010:RLU:3104322.3104425}, batch normalization \cite{DBLP:journals/corr/IoffeS15}, and max-pooling layers. The convolutional layers are with $3 \times 3$ kernel size, and 64 to 1024 channels, depending on the parameter sizes. The second student model is MobileNet, which has a Tensorflow implementation \cite{mobile_tensorflow} on Github. We modified the student models so that the dimension of student model's feature maps equal to the teacher model's. Without further specification, the hyper-parameter $t$ and $\lambda$, as defined in (\ref{eqn:PS}) and (\ref{eqn:KD}), are set to $10$ and $0.1$ respectively. The hyper-parameters $\alpha$, defined in (\ref{eqn:G loss}), is set to 10. The performance metrics of models are the accuracy and the inference time, which is obtained from the average inference time of predicting one CIFAR-100 image 1000 times on one NVIDIA 1080Ti GPU.

\subsection{Teacher Network: DenseNet-40}
This set of experiments uses various CNN models to mimic DenseNet-40.  We compare the results of KDFM with other knowledge distillation methods, and justify the influence of four factors to the accuracy and the inference time: the number of layers, the number of parameters, the value of hyper-parameter $t$ and $\lambda$, defined in (\ref{eqn:PS}) and (\ref{eqn:KD}).

\subsubsection{Comparison with other methods.}
We compared the accuracy of the student network generated by KDFM and other two knowledge distillation methods: logits mimic learning \cite{NIPS2014_5484} and KD \cite{2015arXiv150302531H}.  We also included the results of the model trained without any KD process as the baseline.  The teacher model is DenseNet-40 and the student model has 8 convolutional layers and 8 million trainable parameters. Table \ref{table:3-compare} shows the results of different training methods.  The result indicates that the student model trained by KDFM can acheive similar accuracy as the teacher model's. Logits mimic learning performs poorly. Its accuracy is even lower than that of the baseline in this case.

\begin{table}[]
\begin{center}
\begin{tabular}{lr}
\hline
Method                & Accuracy \\ \hline
Baseline              & 68.53\%  \\
Logits Mimic Learning & 50.95\%  \\
KD                    & 69.14\%  \\
KDFM                 & 74.10\%  \\
Teacher(DenseNet-40)  & 74.23\%  \\ \hline
\end{tabular}
\end{center}
\caption{Testing accuracy for training the student networks with 8 convolutional layers and 8M parameters by Baseline (typical training process), Logits Mimic Learning, KD, and KDFM.}
\label{table:3-compare}
\end{table}

\subsubsection{Different number of layers.}
Table \ref{table:1-simple} summarizes setting of student and teacher models, and their experimental results.  There are four student models which have 2, 4, 6, 8 convolutional layers respectively.  We fixed the number of parameters to 8 millions.  As can be seen, when the number of convolution layers is larger than 4, the student models achieve similar accuracy as the teacher model. Although the model size is not small, the inference time of student models is much shorter than that of the teach network.  Particularly,  the student network with 4 convolution layers has better accuracy than the teacher model, and its inference time is only half of the teacher model's.  

\begin{figure}
\centering
\includegraphics[width=10cm]{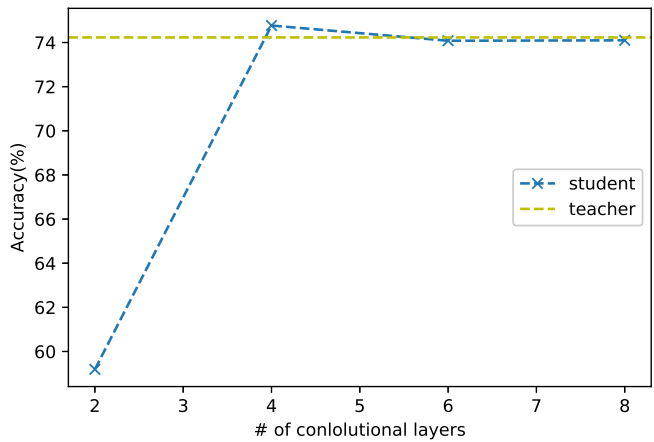}
\caption{Accuracy of student networks with different convolution layers and 8 million parameters, the horizontal line is the accuracy of the teacher network, DenseNet-40.}
\label{fig:layers}
\end{figure}

Figure \ref{fig:layers} plots the accuracy of student networks for different number of layers.  A clear trend is that when the number of layers is larger than 4, the student model can achieve similar accuracy as the teacher model.  However, when the number of layers is small, even with a large number of parameters, the student model cannot learn well as the teacher model.  This result matches the conclusion made in \cite{2016arXiv160305691U}.

\begin{table}[]
\begin{center}
\begin{tabular}{lcrr}
\hline
Model       & No. Parameters & Accuracy & Inference time \\ \hline
2 conv       & $\sim8$M      & 59.19\%  & 3.65ms        \\
4 conv       & $\sim8$M      & 74.77\%  & 2.46ms         \\
6 conv       & $\sim8$M      & 74.08\%  & 2.59ms         \\
8 conv       & $\sim8$M      & 74.10\%  & 2.75ms         \\
DenseNet-40(Teacher) & 1.1M      & 74.23\%  & 5.28ms         \\ \hline
\end{tabular}
\end{center}
\caption{Testing accuracy and inference time for the student networks with 2, 4, 6, and 8 convolutional layers mimicking DenseNet-40 by KDFM.}
\label{table:1-simple}
\end{table}

\subsubsection{Different number of parameters.}
\begin{figure}
\centering
\includegraphics[width=10cm]{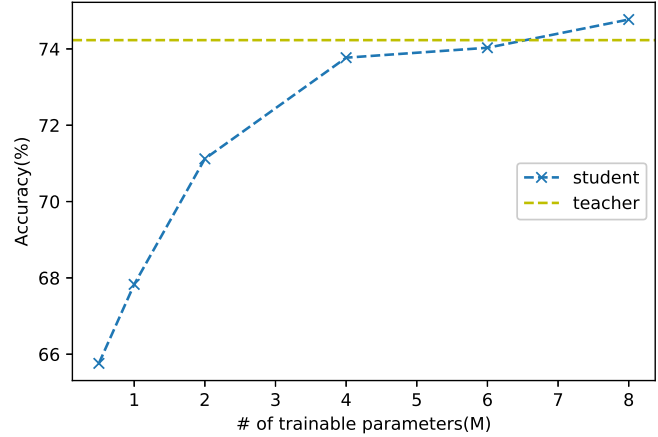}
\caption{Accuracy of student networks with 4 convolution layers and different number of parameters, the horizontal line is the accuracy of the teacher network, DenseNet-40.}
\label{fig:params}
\end{figure}

Table \ref{table:2-more} lists the setting and the results of six student models with different number of parameters.  The number of layers of CNNs is fixed at 4, and the number of parameters are varied from 0.5M, 1M, 2M, 4M, 6M, to 8M.  As can be seen, the more parameters, the better accuracy of the model.  When the number of parameters is larger than or equal to 4M, the accuracy of student model is similar to that of the teacher model.  The difference is less than $1\%$.  Figure \ref{fig:params} shows this trend.

However, the inference time of student models is also increasing as the number of parameter increases.  Nevertheless, even when the number of parameter is 8M, the inference time is still less than half of the teacher model's.  The trade-off between accuracy and the inference time can be used to adjust the student models to fit the requirements of deployments.

\begin{table}[]
\begin{center}
\begin{tabular}{lrrr}
\hline
Model       & No. Parameters & Accuracy & Inference time \\ \hline
4conv-0.5M  & $\sim0.5$M     & 65.76\%  & 1.61ms         \\
4conv-1M    & $\sim1$M       & 67.83\%  & 1.65ms         \\
4conv-2M    & $\sim2$M       & 71.12\%  & 1.73ms         \\
4conv-4M    & $\sim4$M       & 73.77\%  & 2.01ms         \\
4conv-6M    & $\sim6$M       & 73.84\%  & 2.32ms         \\
4conv-8M    & $\sim8$M       & 74.77\%  & 2.46ms         \\
DenseNet-40(Teacher) & 1.1M      & 74.23\%  & 5.28ms         \\ \hline
\end{tabular}
\end{center}
\caption{Testing accuracy and inference time for the student networks with 4 convolutional layers and different numbers of parameters mimicking DenseNet-40 by KDFM.}
\label{table:2-more}
\end{table}

\subsubsection{Different hyper-parameter $t$.}
We validated the influence of hyper-parameter $t$, defined in (\ref{eqn:PS}), to the accuracy of student models.  Table \ref{table:3-t} and Table \ref{table:3-t-8M} show the results for two models, one is a 4 layer CNNs with 2M parameters (small model), and the other is a 4 layer CNNs with 8M parameters (large model).  As can be seen, the best result occurs at $t=5$ for the small model and at $t=10$ for the large model. This phenomenon can be reasoned as follows. When $t$ is small, the soft target does not have enough relaxation to encourage student networks learning the teacher model.  On the other hand, when $t$ is too large, the teacher model losses the disciplines to coach the student models.  For weaker models, smaller $t$ can usually give better accuracy, because they need clearer guidelines to learn.

\begin{table}[]
\begin{center}
\begin{tabular}{lcr}
\hline
Model                & $t$       & Accuracy \\ \hline
4conv with $t=2$     & 2         & 70.14\%  \\
4conv with $t=5$     & 5         & 71.48\%  \\
4conv with $t=10$    & 10        & 71.12\%  \\
4conv with $t=50$    & 50        & 67.35\%  \\
4conv with $t=100$   & 100       & 67.51\%  \\
DenseNet-40(Teacher) & -         & 74.23\% \\ \hline
\end{tabular}
\end{center}
\caption{Testing accuracy for the student networks with 4 convolutional layers, 2M parameters, and different hyper-paramter $t$ mimicking DenseNet-40 by KDFM.}
\label{table:3-t}
\end{table}

\begin{table}[]
\begin{center}
\begin{tabular}{lcr}
\hline
Model                & $t$       & Accuracy \\ \hline
4conv with $t=2$     & 2         & 73.11\%  \\
4conv with $t=5$     & 5         & 74.07\%  \\
4conv with $t=10$    & 10        & 74.77\%  \\
4conv with $t=50$    & 50        & 71.44\%  \\
4conv with $t=100$   & 100       & 70.72\%  \\
DenseNet-40(Teacher) & -         & 74.23\% \\ \hline
\end{tabular}
\end{center}
\caption{Testing accuracy for the student networks with 4 convolutional layers, 8M parameters, and different hyper-paramter $t$ mimicking DenseNet-40 by KDFM.}
\label{table:3-t-8M}
\end{table}

\subsubsection{Different hyper-parameter $\lambda$.}
This experiment compares the accuracy of student models for different hyper-parameter $\lambda$, defined in (\ref{eqn:KD}).  Table \ref{table:3-lambda} and Table \ref{table:3-lambda-8M} show the results for two models, one is a 4 layer CNNs with 2M parameters (small model), and the other is a 4 layer CNNs with 8M parameters (large model).  For both models, the best result occurs at $\lambda=0.1$.  

The results indicate the importance of soft target in knowledge distillation.  For $\lambda=0.1$, the value of soft target dominates the loss function of KD.  This shows that with more information, student models can learn better.  However, if $\lambda$ is set to 0, the information of hard target totally disappears, and the student model cannot learn the best results from the teacher model.

\begin{table}[]
\begin{center}
\begin{tabular}{lcr}
\hline
Model                & $\lambda$       & Accuracy \\ \hline
4conv with $\lambda=0$     & 0           & 70.87\%  \\
4conv with $\lambda=0.1$     & 0.1         & 71.12\%  \\
4conv with $\lambda=0.4$     & 0.4         & 67.30\%  \\
4conv with $\lambda=0.7$    &  0.7        & 66.96\%  \\
DenseNet-40(Teacher) & -         & 74.23\% \\ \hline
\end{tabular}
\end{center}
\caption{Testing accuracy for the student networks with 4 convolutional layers, 2M parameters, and different hyper-parameter $\lambda$ mimicking DenseNet-40 by KDFM.}
\label{table:3-lambda}
\end{table}

\begin{table}[]
\begin{center}
\begin{tabular}{lcr}
\hline
Model      & $\lambda$       & Accuracy \\ \hline
4conv with $\lambda=0$  & 0       & 74.11\% \\
4conv with $\lambda=0.1$     & 0.1     & 74.77\%  \\
4conv with $\lambda=0.4$     & 0.4     & 73.18\%  \\
4conv with $\lambda=0.7$    &  0.7     & 71.68\%  \\
DenseNet-40(Teacher) & -         & 74.23\% \\ \hline
\end{tabular}
\end{center}
\caption{Testing accuracy for the student networks with 4 convolutional layers, 8M parameters, and different hyper-parameter $\lambda$ mimicking DenseNet-40 by KDFM.}
\label{table:3-lambda-8M}
\end{table}

\subsubsection{Different hyper-parameter $\alpha.$}
The hyper-parameter $\alpha$ controls the ratio of GAN and KD in generator's loss function, $L_G=L_{advG} +\alpha L_{KD}$. 
Table \ref{table:alpha} lists the achieved accuracy of student model for different $\alpha$.  The best result occurs at $\alpha=10$ in our experiments.  If we take off GAN $L_{advG}$, as shown in the third line, the accuracy also declines.  

\begin{table}[]
\begin{center}
\begin{tabular}{lcr}
\hline
Model                & $\alpha$       & Accuracy \\ \hline
4conv with $\alpha=1$     & 1         & 69.58\%  \\
4conv with $\alpha=10$    & 10        & 70.62\%  \\
4conv without $L_{advG}$  &  $L_G=L_{KD}$       & 69.57\%  \\
DenseNet-40(Teacher) & -              & 74.23\% \\ \hline
\end{tabular}
\end{center}
\caption{Testing accuracy for the student networks with 4 convolutional layers, 6M parameters, and different hyper-parameter $\alpha$ mimicking DenseNet-40 by KDFM.}
\label{table:alpha}
\end{table}

\subsection{Teacher Network: DenseNet-100}
Since one of the goals for knowledge distillation is to create models easy to deploy on small devices, in this experiment, we used MobileNet (student network) to mimic DenseNet-100 (teacher network). For comparison, we included the results of two other CNNs trained by KDFM.  Both CNNs have 8 convolutional layers, and one has 20.2M parameters; the other has 28.1M parameters.  In addition, the result of MobileNet, trained directly without KDFM, is also included as the baseline.

Table \ref{table:4-dense100} summarizes the results.  The first three rows are the networks trained by KDFM, the fourth row is the result for MobileNet without KD, and the last row is the result of DenseNet-100.   As shown in the first two rows, simple CNNs, even with large amount of parameters, cannot achieve good accuracy as the teacher model.  But their inference times (4.56ms and 5.7ms) are much shorter than that of the original DenseNet-100 (18.02ms).  

The MobileNet trained by KDFM, as shown in the third row, has the best result in terms of model size, accuracy, and inference time.  The number of parameters of MobileNet (3.5M) is less than half of DenseNet-100's (7.2M), and the inference time (2.79ms) is about 6 times faster than the original DenseNet-100 (18.02ms).  Comparing to the baseline, MobileNet without KD, the MobileNet trained by KDFM can achieve 77.20\% accuracy, which is close to that of DenseNet-100 (77.94\%).   

\begin{table}[]
\begin{center}
\begin{tabular}{lrrr}
\hline
Model        & No.Parameters & Accuracy & Inference time \\ \hline
8 conv-20M (KDFM)  & 20.2M      & 74.36\%  & 4.56ms         \\
8 conv-28M (KDFM)  & 28.1M      & 75.25\%  & 5.7ms          \\
MobileNet (KDFM)  & 3.5M       & 77.20\%  & 2.79ms         \\
MobileNet(Baseline) & 3.5M & 72.99\% & 2.79ms          \\
DenseNet-100(Teacher) & 7.2M       & 77.94\%  & 18.02ms       \\ \hline
\end{tabular}
\end{center}
\caption{Testing accuracy and inference time for training simple CNNs with 8 convolutional layers and 20.2M, 28.1M parameters, and MobileNet as student networks by KDFM.}
\label{table:4-dense100}
\end{table}

\subsection{Teacher Network: CondenseNet}
We use CondenseNet-86 \cite{Condense} (with stages [14, 14, 14] and growth [8, 16, 32]) as the teacher network and a smaller CondenseNet-86 (with stages [14, 14, 14] and growth [8, 16, 16]) as the student network using CIFAR-100 dataset.  The results are shown in Table \ref{table:1-condense}. The model trained by KDFM is improved.
\begin{table}[]
\begin{center}
\begin{tabular}{lrrr}
\hline
Model       & No. Parameters & FLOPs    & Accuracy \\ \hline
Smaller CondenseNet-86 (Baseline) & 0.29M & 49.95M & 74.13\% \\
Smaller CondenseNet-86 (KDFM) & 0.29M & 49.95M & 75.01\% \\
CondenseNet-86(Teacher) & 0.55M      & 65.85M  & 76.02\% \\ \hline
\end{tabular}
\end{center}
\caption{Testing accuracy for the smaller CondenseNet-86 mimicking CondenseNet-86 by KDFM and Baseline (typical training process) on CIFAR-100 dataset.}
\label{table:1-condense}
\end{table}

\subsection{ImageNet dataset}
We used MobileNet v2 as the student model to mimic the pre-trained ResNet-152 using ImageNet dataset.  Table \ref{table:3-lambda} shows the experimental results of the testing accuracy using different training methods for the student model.  As can be seen, the baseline method (without KD) can only achieve 68.41\% accuracy.  The KDFM model has the best result, 71.82\%.

\begin{table}[]
\begin{center}
\begin{tabular}{lcrrr}
\hline
Model               & Accuracy & Inference time & FLOPs\\ \hline
MobileNet v2(KDFM)        & 71.82\% & 6ms & 300M\\
MobileNet v2(KD)        & 70.16\% & 6ms & 300M\\
MobileNet v2(KDFM without $L_{advG}$   \&  $L_G=L_{KD}$ )        & 71.32\% & 6ms & 300M\\
MobileNet v2(Baseline)        & 68.01\% & 6ms & 300M\\
ResNet-152(Pre-trained teacher)          & 78.31\% & 21ms & 11G\\ \hline
\end{tabular}
\end{center}
\caption{Testing accuracy for MobileNet v2 as the student networks, trained by KDFM, KD, KDFM without $L_{advG}$, and MobileNet v2 (baseline), to mimic ResNet-152 by KDFM.}
\label{table:3-lambda}
\end{table}


\section{Conclusion and Future Work}
We presented a novel architecture, KDFM, which utilizes generative adversarial networks to achieve knowledge distillation. The experiments demonstrate that KDFM can use simple convolutional neural networks with shallower layers and larger number of trainable parameters to mimic state-of-the-art complicated networks with comparable accuracy and faster inference time. 

The idea of using generative adversarial networks for knowledge distillation is not limited to the DenseNet or image classification tasks, but can be generalized to other types of networks for different applications.  It is also orthogonal to other model compression methods, which means one can use KDFM to generate a student model and apply model pruning or other compression techniques to further reduce the model size and improve the performance.  Last, what is the best student models to be used in KDFM still requires more investigations.  One good feature of KDFM is that other objectives, such as model size, inference speed, power consumption, fitting specific hardware, can be incorporated into the student model design.

\bibliographystyle{splncs04}
\bibliography{egbib}

\begin{thebibliography}{10}
\providecommand{\url}[1]{\texttt{#1}}
\providecommand{\urlprefix}{URL }
\providecommand{\doi}[1]{https://doi.org/#1}

\bibitem{NIPS2014_5484}
Ba, J., Caruana, R.: Do deep nets really need to be deep? In: Ghahramani, Z.,
  Welling, M., Cortes, C., Lawrence, N.D., Weinberger, K.Q. (eds.) Advances in
  Neural Information Processing Systems 27, pp. 2654--2662. Curran Associates,
  Inc. (2014),
  \url{http://papers.nips.cc/paper/5484-do-deep-nets-really-need-to-be-deep.pdf}

\bibitem{DBLP:journals/corr/BojarskiTDFFGJM16}
Bojarski, M., Testa, D.D., Dworakowski, D., Firner, B., Flepp, B., Goyal, P.,
  Jackel, L.D., Monfort, M., Muller, U., Zhang, J., Zhang, X., Zhao, J., Zieba,
  K.: End to end learning for self-driving cars. CoRR  \textbf{abs/1604.07316}
  (2016), \url{http://arxiv.org/abs/1604.07316}

\bibitem{DBLP:journals/corr/ChenDHSSA16}
Chen, X., Duan, Y., Houthooft, R., Schulman, J., Sutskever, I., Abbeel, P.:
  Infogan: Interpretable representation learning by information maximizing
  generative adversarial nets. CoRR  \textbf{abs/1606.03657} (2016),
  \url{http://arxiv.org/abs/1606.03657}

\bibitem{imagenet_cvpr09}
Deng, J., Dong, W., Socher, R., Li, L.J., Li, K., Fei-Fei, L.: {ImageNet: A
  Large-Scale Hierarchical Image Database}. In: CVPR09 (2009)

\bibitem{DBLP:journals/corr/GirshickDDM13}
Girshick, R.B., Donahue, J., Darrell, T., Malik, J.: Rich feature hierarchies
  for accurate object detection and semantic segmentation. CoRR
  \textbf{abs/1311.2524} (2013), \url{http://arxiv.org/abs/1311.2524}

\bibitem{NIPS2014_5423}
Goodfellow, I., Pouget-Abadie, J., Mirza, M., Xu, B., Warde-Farley, D., Ozair,
  S., Courville, A., Bengio, Y.: Generative adversarial nets. In: Ghahramani,
  Z., Welling, M., Cortes, C., Lawrence, N.D., Weinberger, K.Q. (eds.) Advances
  in Neural Information Processing Systems 27, pp. 2672--2680. Curran
  Associates, Inc. (2014),
  \url{http://papers.nips.cc/paper/5423-generative-adversarial-nets.pdf}

\bibitem{NIPS1992_647}
Hassibi, B., Stork, D.G.: Second order derivatives for network pruning: Optimal
  brain surgeon. In: Hanson, S.J., Cowan, J.D., Giles, C.L. (eds.) Advances in
  Neural Information Processing Systems 5, pp. 164--171. Morgan-Kaufmann
  (1993),
  \url{http://papers.nips.cc/paper/647-second-order-derivatives-for-network-pruning-optimal-brain-surgeon.pdf}

\bibitem{7780459}
He, K., Zhang, X., Ren, S., Sun, J.: Deep residual learning for image
  recognition. In: 2016 IEEE Conference on Computer Vision and Pattern
  Recognition (CVPR). pp. 770--778 (June 2016). \doi{10.1109/CVPR.2016.90}

\bibitem{DBLP:resnet}
He, K., Zhang, X., Ren, S., Sun, J.: Deep residual learning for image
  recognition. CoRR  \textbf{abs/1512.03385} (2015),
  \url{http://arxiv.org/abs/1512.03385}

\bibitem{2015arXiv150302531H}
{Hinton}, G., {Vinyals}, O., {Dean}, J.: {Distilling the Knowledge in a Neural
  Network}. ArXiv e-prints  (Mar 2015)

\bibitem{DBLP:journals/corr/HowardZCKWWAA17}
Howard, A.G., Zhu, M., Chen, B., Kalenichenko, D., Wang, W., Weyand, T.,
  Andreetto, M., Adam, H.: Mobilenets: Efficient convolutional neural networks
  for mobile vision applications. CoRR  \textbf{abs/1704.04861} (2017),
  \url{http://arxiv.org/abs/1704.04861}

\bibitem{8099726}
Huang, G., Liu, Z., v.~d. Maaten, L., Weinberger, K.Q.: Densely connected
  convolutional networks. In: 2017 IEEE Conference on Computer Vision and
  Pattern Recognition (CVPR). pp. 2261--2269 (July 2017).
  \doi{10.1109/CVPR.2017.243}

\bibitem{Condense}
Huang, G., Liu, S., van~der Maaten, L., Weinberger, K.Q.: Condensenet: An
  efficient densenet using learned group convolutions. In: The IEEE Conference
  on Computer Vision and Pattern Recognition (CVPR) (June 2018)

\bibitem{DBLP:journals/corr/HubaraCSEB16}
Hubara, I., Courbariaux, M., Soudry, D., El{-}Yaniv, R., Bengio, Y.: Quantized
  neural networks: Training neural networks with low precision weights and
  activations. CoRR  \textbf{abs/1609.07061} (2016),
  \url{http://arxiv.org/abs/1609.07061}

\bibitem{DBLP:journals/corr/IoffeS15}
Ioffe, S., Szegedy, C.: Batch normalization: Accelerating deep network training
  by reducing internal covariate shift. CoRR  \textbf{abs/1502.03167} (2015),
  \url{http://arxiv.org/abs/1502.03167}

\bibitem{vision_networks}
Khlestov, I.: vision\_networks.
  \url{https://github.com/ikhlestov/vision_networks} (2017)

\bibitem{cifar100}
Krizhevsky, A., Hinton, G.: Learning multiple layers of features from tiny
  images  \textbf{1} (01 2009)

\bibitem{Krizhevsky:2012:ICD:2999134.2999257}
Krizhevsky, A., Sutskever, I., Hinton, G.E.: Imagenet classification with deep
  convolutional neural networks. In: Proceedings of the 25th International
  Conference on Neural Information Processing Systems - Volume 1. pp.
  1097--1105. NIPS'12, Curran Associates Inc., USA (2012),
  \url{http://dl.acm.org/citation.cfm?id=2999134.2999257}

\bibitem{8099502}
Ledig, C., Theis, L., Huszár, F., Caballero, J., Cunningham, A., Acosta, A.,
  Aitken, A., Tejani, A., Totz, J., Wang, Z., Shi, W.: Photo-realistic single
  image super-resolution using a generative adversarial network. In: 2017 IEEE
  Conference on Computer Vision and Pattern Recognition (CVPR). pp. 105--114
  (July 2017). \doi{10.1109/CVPR.2017.19}

\bibitem{8237566}
Mao, X., Li, Q., Xie, H., Lau, R.Y.K., Wang, Z., Smolley, S.P.: Least squares
  generative adversarial networks. In: 2017 IEEE International Conference on
  Computer Vision (ICCV). pp. 2813--2821 (Oct 2017).
  \doi{10.1109/ICCV.2017.304}

\bibitem{DBLP:journals/corr/MirzaO14}
Mirza, M., Osindero, S.: Conditional generative adversarial nets. CoRR
  \textbf{abs/1411.1784} (2014), \url{http://arxiv.org/abs/1411.1784}

\bibitem{6467091}
Mizotin, M., Benois-Pineau, J., Allard, M., Catheline, G.: Feature-based brain
  mri retrieval for alzheimer disease diagnosis. In: 2012 19th IEEE
  International Conference on Image Processing. pp. 1241--1244 (Sept 2012).
  \doi{10.1109/ICIP.2012.6467091}

\bibitem{Nair:2010:RLU:3104322.3104425}
Nair, V., Hinton, G.E.: Rectified linear units improve restricted boltzmann
  machines. In: Proceedings of the 27th International Conference on
  International Conference on Machine Learning. pp. 807--814. ICML'10,
  Omnipress, USA (2010),
  \url{http://dl.acm.org/citation.cfm?id=3104322.3104425}

\bibitem{2016arXiv160601583O}
{Odena}, A.: {Semi-Supervised Learning with Generative Adversarial Networks}.
  ArXiv e-prints  (Jun 2016)

\bibitem{pmlr-v70-odena17a}
Odena, A., Olah, C., Shlens, J.: Conditional image synthesis with auxiliary
  classifier {GAN}s. In: Precup, D., Teh, Y.W. (eds.) Proceedings of the 34th
  International Conference on Machine Learning. Proceedings of Machine Learning
  Research, vol.~70, pp. 2642--2651. PMLR, International Convention Centre,
  Sydney, Australia (06--11 Aug 2017),
  \url{http://proceedings.mlr.press/v70/odena17a.html}

\bibitem{pmlr-v48-reed16}
Reed, S., Akata, Z., Yan, X., Logeswaran, L., Schiele, B., Lee, H.: Generative
  adversarial text to image synthesis. In: Balcan, M.F., Weinberger, K.Q.
  (eds.) Proceedings of The 33rd International Conference on Machine Learning.
  Proceedings of Machine Learning Research, vol.~48, pp. 1060--1069. PMLR, New
  York, New York, USA (20--22 Jun 2016),
  \url{http://proceedings.mlr.press/v48/reed16.html}

\bibitem{Romero15-iclr}
Romero, A., Ballas, N., Kahou, S.E., Chassang, A., Gatta, C., Bengio, Y.:
  Fitnets: Hints for thin deep nets. In: In Proceedings of ICLR (2015)

\bibitem{NIPS2016_6125}
Salimans, T., Goodfellow, I., Zaremba, W., Cheung, V., Radford, A., Chen, X.,
  Chen, X.: Improved techniques for training gans. In: Lee, D.D., Sugiyama, M.,
  Luxburg, U.V., Guyon, I., Garnett, R. (eds.) Advances in Neural Information
  Processing Systems 29, pp. 2234--2242. Curran Associates, Inc. (2016),
  \url{http://papers.nips.cc/paper/6125-improved-techniques-for-training-gans.pdf}

\bibitem{DBLP:MBv2}
Sandler, M., Howard, A.G., Zhu, M., Zhmoginov, A., Chen, L.: Inverted residuals
  and linear bottlenecks: Mobile networks for classification, detection and
  segmentation. CoRR  \textbf{abs/1801.04381} (2018),
  \url{http://arxiv.org/abs/1801.04381}

\bibitem{mobile_tensorflow}
Shi, Z.: Mobilenet. \url{https://github.com/Zehaos/MobileNet} (2017)

\bibitem{DBLP:journals/corr/SimonyanZ14a}
Simonyan, K., Zisserman, A.: Very deep convolutional networks for large-scale
  image recognition. CoRR  \textbf{abs/1409.1556} (2014),
  \url{http://arxiv.org/abs/1409.1556}

\bibitem{intelligent-manu}
Teti, R., Kumara, S.: Intelligent computing methods for manufacturing systems
  \textbf{46},  629--652 (12 1997)

\bibitem{2016arXiv160305691U}
{Urban}, G., {Geras}, K.J., {Ebrahimi Kahou}, S., {Aslan}, O., {Wang}, S.,
  {Caruana}, R., {Mohamed}, A., {Philipose}, M., {Richardson}, M.: {Do Deep
  Convolutional Nets Really Need to be Deep and Convolutional?} ArXiv e-prints
  (Mar 2016)

\bibitem{iclr2018training}
Xu, Z., Hsu, Y.C., Huang, J.: Training shallow and thin networks for
  acceleration via knowledge distillation with conditional adversarial networks
  (2018), \url{https://openreview.net/forum?id=BJbtuRRLM}

\end{thebibliography}
\end{document}